\documentclass[conference]{IEEEtran}
\IEEEoverridecommandlockouts
\usepackage{cite}
\usepackage{amsmath,amssymb,amsfonts}
\usepackage{graphicx}
\usepackage{textcomp}
\usepackage{xcolor}
\usepackage{hyperref}
\usepackage{caption}
\usepackage{booktabs}
\usepackage{algorithm}
\usepackage{algpseudocode}
\usepackage{subcaption}
\def\BibTeX{{\rm B\kern-.05em{\sc i\kern-.025em b}\kern-.08em
    T\kern-.1667em\lower.7ex\hbox{E}\kern-.125emX}}

\begin{document}

\title{Towards a Fully Autonomous UAV Controller for Moving Platform Detection and Landing

\thanks{This work has been supported by the European Union’s Horizon 2020 research and innovation programme under grant agreement No 739551 (KIOS CoE) and from the government of the Republic of Cyprus through the Directorate General for European Programmes, Coordination and Development.}
}

\author{\IEEEauthorblockN{Michalis Piponidis, Panayiotis Aristodemou and Theocharis Theocharides}
\IEEEauthorblockA{\textit{KIOS Research and Innovation Center of Excellence}\\
\textit{Department of Electrical and Computer Engineering}\\
\textit{University of Cyprus}\\
Nicosia, Cyprus\\
\{mpipon01,parist02,ttheocharides\}@ucy.ac.cy}
}

\maketitle
\begin{abstract}
 While Unmanned Aerial Vehicles (UAVs) are increasingly deployed in several missions, their inability of reliable and consistent autonomous landing poses a major setback for deploying such systems truly autonomously. In this paper we present an autonomous UAV landing system for landing on a moving platform. In contrast to existing attempts, the proposed system relies only on the camera sensor, and has been designed as lightweight as possible. The proposed system can be deployed on a low power platform as part of the drone payload, whilst being indifferent to any external communication or any other sensors. The system relies on a Neural Network (NN) based controller, for which a target and environment agnostic simulator was created, used in training and testing of the proposed system, via Reinforcement Learning (RL) and Proximal Policy Optimization (PPO) to optimally control and steer the drone towards landing on the target. Through real-world testing, the system was evaluated with an average deviation of 15cm from the center of the target, for 40 landing attempts.
\end{abstract}

\begin{IEEEkeywords}
UAVs, drone, autonomous landing, machine learning, neural networks, computer vision, object detection
\end{IEEEkeywords}

\section{Introduction}
With the continuous growth of Unmanned Aerial Vehicles (UAVs) usage in many applications, their ability to be as autonomous as possible is becoming a primary objective. Multi-rotor UAVs especially are extremely useful due to their high maneuverability in various terrains and are being adopted in numerous missions such as search and rescue \cite{searchandrescue}, military and tracking operations \cite{military} and many more. Autonomous UAV landing, especially on a moving target is one of the most vital and simultaneously challenging operations of a UAV to achieve full autonomy. Research in this area is still lacking and human operators are still necessary which limits the countless possibilities of UAVs, that range from rapid recovery of a fleet \cite{recovery} to the utilization of mobile recharging stations \cite{recharge} where UAVs can autonomously recharge during missions.

In the past few years, UAVs are becoming eminently dependent on external communication signals, namely wireless communication and Global Positioning Systems (GPS). While such systems provide UAVs with extremely reliable and safe operational abilities, they become vulnerable to various attacks such as signal jamming and spoofing, which can cause poor and false readings or even accidents \cite{gpsattacks}. Additionally, poor GPS signal can cause unacceptable amounts of deviation reaching up to 4m which eliminates the likelihood of a reliable system that can be used repeatably without any problems \cite{gpsunstable}. Furthermore, in search and rescue operations, and emergency response scenarios, where UAVs have been adopted to provide first responders with extremely valuable and vital information, the availability of communication networks or GPS may not be readily available. Near-field communication (NFC) that has also been proposed as a means of communication between the UAV and the landing platform is impractical in such scenarios, as the UAV still has to detect and move towards the platform for NFC to be effective.

In this paper, we therefore propose, a lightweight, real-time, neural-network based landing controller that relies only on its own sensors and controls the UAV towards a moving target and autonomously lands it. The contribution of this work is a complete system designed and optimized from the very beginning to be deployed on a UAV. The system is evaluated on an NVIDIA\textsuperscript{\textregistered} Jetson Xavier\textsuperscript{TM} NX embedded platform and a DJI Mavic Air UAV. In particular, this work advances the state of the art in the following ways:
\begin{itemize}
  \item The proposed system does not require any external communication signals between the UAV and the target, as it relies only on the on-board UAV camera sensor.
  \item The controller utilizes an accurate target detector and extracts a positional relationship between the UAV and the target platform.
  \item We developed a target- and environment-agnostic simulator to generate training data, used to train the controller under different conditions, in a consistent and fast manner.
  \item The controller is based on a Neural Network (NN), trained using Reinforcement Learning (RL) that uses the positional relationship between the target and the UAV as its input, and accordingly outputs the optimal UAV control signals. Further, we perform design exploration and optimization to reduce its size and latency while keeping the accuracy at desirable levels.
  \item We evaluated and validated the proposed controller, through a set of real-world experimental scenarios that yielded a 15cm average deviation from the center of the platform for 40 landing experiments. The controller's resource requirements were also validated by deploying it on an embedded platform and extracting performance metrics.
\end{itemize}

The rest of the paper is structured as follows: Section \ref{related} discusses the related work that has been done on this subject. Section \ref{methodology} explains the modules developed in the proposed framework. Section \ref{experimental} describes the experimental platform that is used for evaluating the system and shows the experimental results. Finally, in Section \ref{conclusions} we conclude our work and discuss possible future work.

\section{Related Work} \label{related}

UAV autonomous landing systems has been a hot topic for researchers, especially since the advancements in the UAV industry that happened in the last decade. Initial works focused on detecting the landing platform, based on traditional digital image processing methods. Xiaoyun et al. \cite{huang} designed a color marker platform that gets detected during the landing process by exploiting its colors (red, green and blue), which makes it easy to detect by the combined approach of vanishing point of parallel lines and Levenberg-Marquardt (L-M) optimization method. A similar approach is shown in \cite{chen}, where the target is detected using adaptive thresholding. Ramos et al. \cite{ramos} use a landing platform with the Helipad design and using quadrilateral object detection with a Feature Extraction module based on Histogram of Oriented Gradients (HOG) and a Supervised Learning Classiﬁer are able to detect it reliably. Demirhan et al. \cite{demirhan} exploit the contour properties of the Helipad design to detect the platform. Although this approach has very good performance, it can only reliably detect the platform from a maximum altitude of 165cm, which makes it impractical for typical real-world applications. Many research works use the UAV's GPS to assist the landing procedure \cite{saad} \cite{huang} \cite{liang}. The GPS accuracy usually ranges between 2-3m \cite{janousek}, which, depending on the size of the landing platform, may require additional information or human control for reliable landing. The usage of Aruco markers \cite{armarkers} is another popular choice for detecting the landing platform, as seen in \cite{vankadari} \cite{wubben} \cite{zhao} \cite{araar}. Although they are very accurate and efficient in low altitudes where the marker is clearly visible, when the UAV is higher ($>$10m), they are harder to detect as we saw from our experiments. This is caused by the combination of their complex design and low resolution. A different approach is presented in \cite{patruno}, where the system is able to ensure the UAV pose estimation at different altitude levels by a coarse-to-fine approach, detecting a different part of the platform at each level. A mathematical approach based on the inclination angle and state of the UAV is also presented in \cite{alijani}, however it was only evaluated by simulation and the platform is assumed to move at a constant velocity.

The use of Machine Learning (ML) in designing landing controllers for multi-rotor UAVs, is also relatively new, as such controllers are typically complex and resource-constrained, energy-aware NN-based controllers only recently gained popularity through various edge-based optimizations \cite{he}. Thus, RL based controllers, that use Least Square Policy Iteration (LSPI) to learn the optimal UAV control policies have been proposed \cite{vankadari} \cite{shaker}, all of which however use static landing platforms. Moreover, Ramos et al. \cite{ramos2} demonstrate a Deep Deterministic Policy Gradients (DDPG) RL algorithm that can achieve really good accuracy when the positions of the UAV and the moving platform are known. These works prove that RL based controllers are one of the best approaches for solving this problem.

Motivated by the need of a fully autonomous system, in contrast to existing works, we design our controller to use only on-board sensing and computational resources, targeting landing on a moving platform, without constraints on the platform's location or movement patterns (i.e. the platform can move freely with variable direction and/or speed). It is also important to note that the evaluation methods and metrics are not standardized yet - this is due to the fact that these systems are fairly complex, so they can be evaluated using many different metrics. The most descriptive and complete comparison scheme we found is shown in \cite{wubben}, which compares state of the art frameworks in regards of their \textit{Accuracy} (i.e. the average deviation from the center of the platform), the UAV's \textit{Landing Speed}, the \textit{Maximum Altitude} from which it can work and whether it works with a \textit{Moving Platform} and \textit{Outdoors}. We thus use these metrics to evaluate our proposed controller.

\section{Methodology} \label{methodology}

Our system consists of the following modules:
\begin{itemize}
  \item \textit{Object detector:} reliable and fast detector in order to detect the position of the landing platform in real-time.
  \item \textit{Target positioning:} extracts various information providing the positional relationship between the target and UAV.
  \item \textit{Simulator:} virtual environment where a UAV can be simulated, generating the controller's training and evaluation data.
  \item \textit{RL based controller:} trained NN for taking decisions about the UAV movement, based on the positional information.
\end{itemize}
Each of these modules was designed targeting a resource-constrained embedded implementation requirements. We aim for the lowest possible resources such as memory footprint and ultra-low power consumption, while still being able to successfully complete the landing task with the required performance and accuracy.

\begin{figure}[!tb]
\centering
\subfloat[Frame Capture \& Processing]{\includegraphics[width=1.2in]{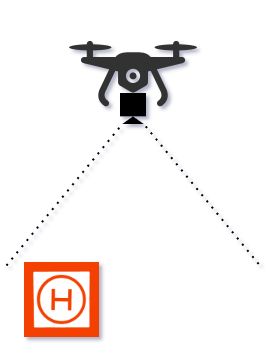} \label{framecapture}}
\hfil
\subfloat[UAV Movement]{\includegraphics[width=0.7in]{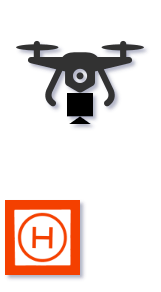} \label{dronemovement}}
\hfil
\subfloat[Repeat]{\includegraphics[width=0.65in]{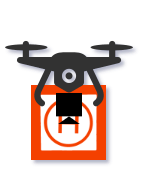} \label{repeat}}
\caption{System visual representation}
\label{systemvisualrepresentation}
\end{figure}

\begin{figure}[!bt]
\centering
\includegraphics[width=2.65in]{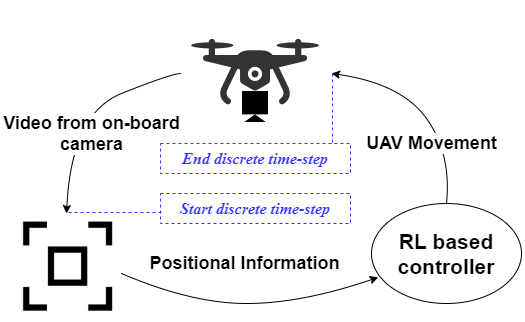}
\caption{System communication diagram}
\label{systemcommunication}
\end{figure}

A visual representation of our system landing process is shown in \figurename \ref{systemvisualrepresentation}, while the time and data flow between the modules can be seen in \figurename \ref{systemcommunication}. We partition the process in discrete time-steps. In each time-step, the process starts with the detector receiving the camera feed. The detector then finds the position of the platform in each frame, for which the information about the positional relationship between the target and UAV is extracted (\figurename \ref{framecapture}). That information is sent to the controller, which outputs the optimal UAV movement commands (\figurename \ref{dronemovement}). This process is repeated in every time-step, until the UAV lands on the platform (\figurename \ref{repeat}). The duration of each time-step is defined in processed Frames Per Second (FPS), encapsulating all actions as shown in \figurename \ref{systemcommunication}.

\subsection{Target Detection and Parameter Estimation}

We utilize a red-colored helipad style design as shown in \figurename \ref{systemvisualrepresentation} for the targeted platform. The reasoning behind the chosen shape and color, is because it is simple and can be distinguished by the detector easier in higher altitudes. Furthermore, the color red is quite rare in nature meaning that the possibility of a false positive detection is significantly decreased.

Given the abundance of object detectors available, we decided to use a state of the art detector, optimized for UAVs, DroNet \cite{dronet}. However, our system is vision-agnostic and any detection method (e.g. \cite{unel}, \cite{jiang}, \cite{zhang}) can be used, depending on the specific application requirements. Following up the time-step approach explained earlier, we only target 8FPS instead of the 30FPS that the camera provides, further reducing the computational strain. To determine how strict this requirement is, we provide a brief analysis next. Let us consider a hovering altitude of 5m; the UAV camera covers $\approx$9x9m of the ground. In order for the moving platform to not be detected in successive frames, it would need to cover 9m in 0.1s. This implies a platform speed of $9*8 = 72\frac{m}{s}$, which is of course quite unnatural, and means that if needed, we can further reduce the processed FPS to lower the system requirements. Even more, UAVs typically fly at altitudes of 20m and above, which implies a much larger (in terms of surface area) view.

To extract information about the positional relationship between the target and the UAV, the detected bounding box was used. The bounding box is essentially the rectangle that surrounds the detected object as provided by the detector; it consists of four coordinates in the image: \textit{xmin}, \textit{ymin}, \textit{xmax} and \textit{ymax}. To extract the information, the first step is to find the center of the bounding box as shown in (\ref{boundingboxcenterpoints}).

\begin{equation}
\begin{aligned}
    &x_c = xmax - \frac{(xmax-xmin)}{2}\\
    &y_c = ymax - \frac{(ymax-ymin)}{2}
\end{aligned}
    \label{boundingboxcenterpoints}
\end{equation}

Where $(x_c,y_c)$ is the center point of the bounding box. Afterwards, the distance of the center point to the image center, in each axis must be found using (\ref{hvpdist}).

\begin{equation}
\begin{aligned}
    &HPDist = x_c - \frac{imageWidth}{2}\\
    &VPDist = y_c - \frac{imageHeight}{2}
    \label{hvpdist}
\end{aligned}
\end{equation}

Where $HPDist$, $VPDist$ are the horizontal and vertical pixel distances from the center point to the horizontal and vertical axis of the image respectively. $imageWidth$, $imageHeight$ are the image width and height in pixels respectively.

The first variable that is calculated is the Horizontal Angle, also known as Rotation Angle, which is the angle of the target with respect to the heading of the UAV (longitudinal axis). To compute the Horizontal Angle, (\ref{horizontalangleequation}) is used.

\begin{equation}
    \theta = \frac{HFOV \cdot HPDist}{imageWidth}
    \label{horizontalangleequation}
\end{equation}

Where $\theta$ is the Horizontal Angle and $HFOV$ is the Horizontal Field of View of the camera in degrees. The key part of this variable is that depending on the sign, we know meticulously how many degrees the target is positioned right or left from the UAV.

The second variable is the Vertical Angle, which is the angle of the target with respect to the UAV's vertical axis. Calculating this variable involves a two step process. Firstly, we use a similar approach as the Horizontal Angle, as shown in (\ref{verticalangleequation1}). 

\begin{equation}
    \phi = \frac{VFOV \cdot VPDist}{imageHeight}
    \label{verticalangleequation1}
\end{equation}

Where $\phi$ is the vertical angle from the UAV camera's gimbal heading in relation to the target and $VFOV$ is the Vertical Field of View of the camera in degrees. The sign of $\phi$ indicates if the target is above or below the gimbal's heading. Given the fact that the gimbal is not static, equation (\ref{verticalangleequation}) is extracted.

\begin{equation}
    \omega = \phi + \alpha
    \label{verticalangleequation}
\end{equation}

Where $\alpha$ is the pitch angle of the gimbal from the UAV's vertical axis and $\omega$ is the Vertical Angle. Vertical Angle is used to dynamically change the gimbal of the UAV to keep the target at the center of the image on the vertical axis.

Another crucial variable is the distance from the UAV to the landing platform. Both Horizontal and Vertical angles are required for accurately computing the distance. By exploiting basic triangulation properties, (\ref{horizontaldistanceequation}) is derived.

\begin{equation}
    d_1 = h \cdot \tan{(\omega)}
    \label{horizontaldistanceequation}
\end{equation}

Where $d_1$ is distance from the UAV to the platform on the UAV's longitudinal axis and $h$ is the height of the UAV. The next step for calculating the distance is (\ref{platformdistanceequation}).

\begin{equation}
    d_2 = \frac{d_1}{\cos{(\theta)}}
    \label{platformdistanceequation}
\end{equation}

Where $d_2$ is the direct distance from the UAV to the landing platform.

Finally, using (\ref{targetvelocityequation}) the platform speed can be calculated.

\begin{equation}
    V_p =\left( \frac{D_2 - D_1}{T} \right) + V_d
    \label{targetvelocityequation}
\end{equation}

Where $D_2$ is the current distance of the platform, $D_1$ is the previous distance, $T$ the time between the two measured distances and $V_d$ is the speed of the UAV.

\subsection{Control Neural Network}

We used Unity's ML-Agents Toolkit \cite{mlagents} to train the UAV controller NN. ML-Agents is a powerful framework that allows intelligent agents to learn through a combination of Deep RL and Imitation Learning. In particular, we used the ML-Agents RL framework with the Proximal Policy Optimization (PPO) algorithm \cite{ppo}.

\begin{figure}[!tb]
  \centering
  \begin{minipage}[b]{0.49\columnwidth} 
    \includegraphics[width=\textwidth]{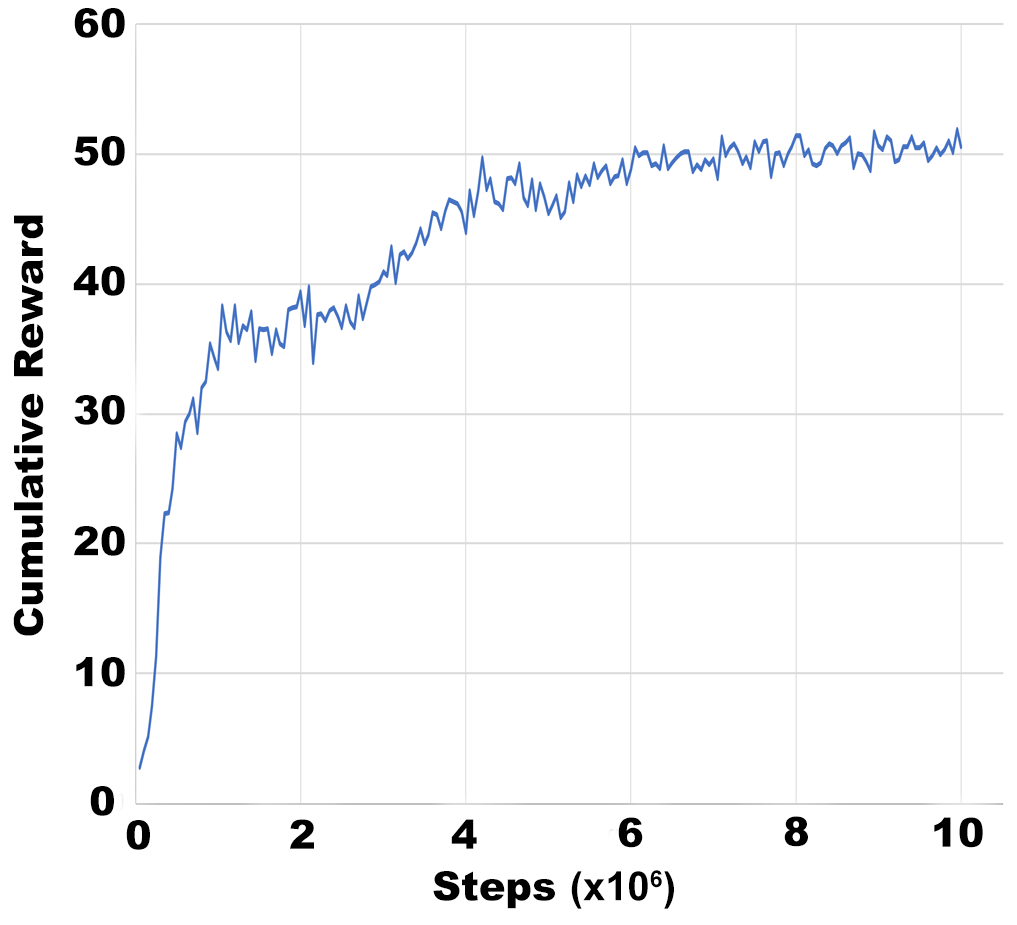}
    \caption{RL Training Results}
    \label{trainingresults}
  \end{minipage}
  \hfill
  \begin{minipage}[b]{0.49\columnwidth}
    \includegraphics[width=\textwidth]{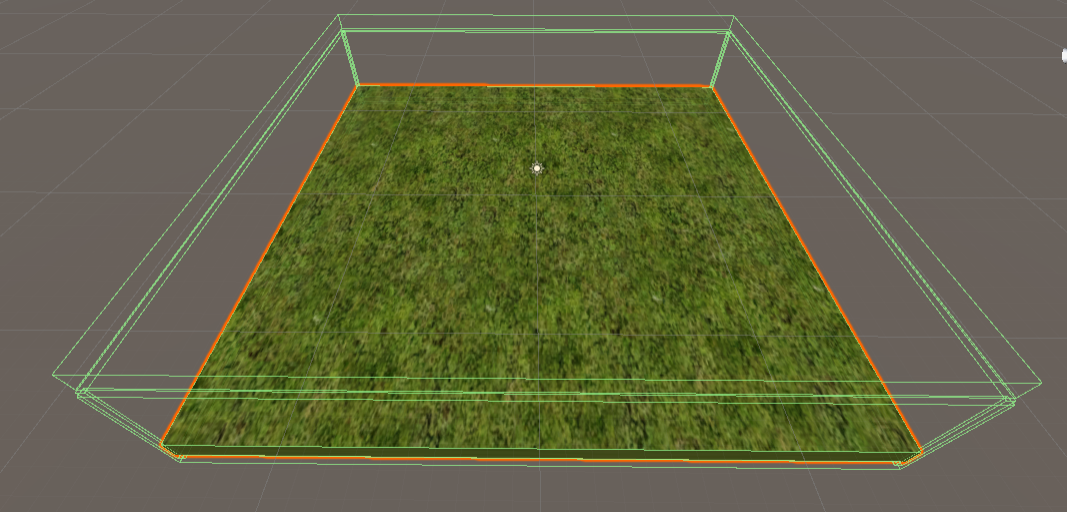}
    \caption{Simulator}
    \label{simulatorlevel}
  \end{minipage}
\end{figure}

The decision making NN receives the information extracted from the detector as the network input, also known as Observations in RL. These include the \textit{Rotation Difference}, \textit{Horizontal Distance} and \textit{Vertical Distance}. These values were normalized because as shown by an ablation study \cite{normalization}, normalized observations contribute to better performance in PPO based RL. The action space has four UAV control outputs: pitch, roll, yaw and throttle. The pseudocode of the reward algorithm is shown in Algorithm \ref{pseudocode}. Negative rewards are larger than positive rewards in order to prevent unnecessary movement while the values have been chosen empirically after various simulations. \figurename \ref{trainingresults} shows that the training process progressed smoothly, with the ﬁnal NN reliably following and landing on the target, using consistent and smooth movements.

\begin{algorithm}
\small
\caption{RL Rewards Pseudocode}\label{pseudocode}
\begin{algorithmic}
\If{UAV turned $10^{\circ}$ closer to the target}
    \State Give $+0.1$ Reward
\ElsIf{UAV turned $10^{\circ}$ further from the target}
    \State Give $-0.3$ Reward
\EndIf
\If{UAV is within $5^{\circ}$ from the target}
    \State Give $+0.1$ Reward
\EndIf
\If{UAV moved horizontally $0.2m$ towards the target}
    \State Give $+0.1$ Reward
\ElsIf{UAV moved horizontally $0.2m$ away from the target}
    \State Give $-0.3$ Reward
\EndIf
\If{UAV moved vertically $0.2m$ towards the target}
    \State Give $+0.1$ Reward
\ElsIf{UAV moved vertically $0.2m$ away from the target}
    \State Give $-0.3$ Reward
\EndIf
\If{UAV entered the landing target}
    \State Give $+15$ Reward
\EndIf
\If{UAV stayed in the landing target}
    \State Give $+0.2$ Reward
\EndIf
\end{algorithmic}
\end{algorithm}

The NN was optimized to reduce it from its original size of 78KB to 7KB, for a total of 91\% size decrease. The computational complexity of the NN is also notably decreased, resulting in faster inference while keeping the desired performance, making it suitable for the intended on-board implementation. Due to space limitations, we omit the optimization details, however, we used standard pruning and quantization approaches.

\subsection{Training Simulator}

The simulator provides a controlled and customizable environment so that any environmental data can be used to train the controller, while also allowing the control of several parameters, such as the UAV and platform positions and velocities. Moreover, it provides a simulated environment where we can perform our training without the need of deploying an actual UAV, which can be time consuming, expensive and can result in accidents. The primary objective of the simulator is to serve as the environment where the RL based controller will be trained, via various scenarios as stated previously. This greatly improves the training process as very large numbers of varied training data can be easily generated, which results in better training and avoids over-fitting. The environment is simulated realistically based on real-world data, thus it can be assured that training data are valid and applicable to real-world situations. The simulator is built using the Unity game engine \cite{unity}. The virtual environment (\figurename \ref{simulatorlevel}) consists of a large hollow rectangular prism as the level where the UAV can fly and a landing platform with adjustable movement parameters. The virtual UAV \cite{virtualdrone} that was used provides a realistic model that moves by adding force to each propeller and uses PID controllers for stabilization.

\section{Experimental Setup and Results} \label{experimental}

While the proposed controller is indented to be deployed directly on the UAV, for testing and troubleshooting purposes, it was running on a laptop, communicating with the UAV via a video streaming server and an interface built using Robotic Operating System (ROS\textsuperscript{TM}) \cite{ros} for transmitting the controller's output signals back to the UAV. We detail our experimental framework next. 

\subsection{Experimental Setup}

A labeled dataset consisting of 3419 real-world images in various altitudes, angles and lighting conditions (using a UAV capturing photos of the platform) was used to train the detector. We trained DroNet through Darknet \cite{darknet}, reaching 85.9\% mean Average Precision (mAP). We observed extremely high accuracy (more than 95\%) detecting the target for altitudes up to 20m, with greater altitudes also resulting in high accuracy as well (more than 80\%).

We used the DJI Mavic Air UAV because it provides the necessary features required for the application and an easy to use Software Development Kit (SDK). Furthermore, it has a 3-axis camera gimbal utilized by the detector and positional relationship extraction. We kept the UAV velocity constant at 0.4$\frac{m}{s}$ for evaluation purposes. We also developed an Android application using the Mobile SDK, providing full control over the UAV's movement and gimbal angles, which we used to translate the movement commands issued by the controller NN into UAV movement commands, while the gimbal angle was adjusted accordingly after each time-step. We used ROS to communicate between the modules and a laptop with Intel\textsuperscript{\textregistered} Core\textsuperscript{TM} i7-6500U 2.5GHz and 16GB DDR3 RAM running Ubuntu 20.04 LTS. The controller thus receives the UAV's camera stream, performs the target detection and positional extraction, which are then used by the NN controller, which outputs the control signals. These are then transmitted to the UAV using the ROS interface. Further, we built an 100x100cm landing platform consisting of the 60x60cm target in the middle and 20cm padding used as a safety measure, in case the UAV lands on the edge of the platform. The platform was moved using a programmable custom-build robotic platform that enabled us to create various speeds and movements and generate random paths on the ground.

\subsection{Experimental Results}

We used the DJI Mavic Air UAV, the controller (as described earlier) and the custom-built moving landing platform, and performed a total of 40 test runs starting from various altitudes and under variable landing platform speeds and directions. The results are shown in Table \ref{table:movingplatformtesting}. The evaluation was based strictly on the detector outputs, thus any associated limitations and inaccuracies of the vision algorithm are also considered. From the results we can see that 30 out of 40 runs landed inside the 60x60cm target, which yields an accuracy of 75\%. Moreover, the average distances were measured from the center of the landing platform to the UAV at its landing position, yielding an average deviation of $\approx$15cm.

\begin{table}[!tb]
\renewcommand{\arraystretch}{1.3} 
\caption{Moving platform testing results}
\label{table:movingplatformtesting}
\begin{center}
\begin{tabular}{ c  c  c } 
\toprule
\textbf{Runs} & \textbf{Average Distance from the} & \textbf{Landings inside}\\
& \textbf{center of the platform} & \textbf{the target}\\
\midrule
\textit{1-5} & $20cm$ & $3$ \\ 
\textit{6-10} & $4cm$ & $5$ \\
\textit{11-15} & $35cm$ & $2$ \\ 
\textit{16-20} & $10cm$ & $4$ \\ 
\textit{21-25} & $21cm$ & $3$ \\ 
\textit{26-30} & $8cm$ & $5$ \\ 
\textit{31-35} & $15cm$ & $3$ \\ 
\textit{36-40} & $12cm$ & $5$ \\ 
\bottomrule
\end{tabular}
\end{center}
\end{table}

Table \ref{comparison} summarizes the results and compares them to related work shown in Section \ref{related}. We can observe that our work is the only one that works with a moving target outdoors, which in itself is a major step forward towards deployment in practical applications. Also, our evaluation metrics are comparable to similar state of the art works. It should be taken into consideration that these metrics are based on the laptop implementation of the system. We anticipate that on-board deployment, will significantly improve performance due to the absence of the communication overheads between the camera and the controller. To verify this, we implemented the detector and the NN controller on an NVIDIA\textsuperscript{\textregistered} Jetson Xavier\textsuperscript{TM} NX embedded platform, which we use as computational payload, to demonstrate its performance. The detector used $\approx$1.3GB of RAM, while the controller used used $\approx$450MB of RAM, for a total of 25\% of the available memory. The detector achieved an average speed of 20FPS (including the positional relationship information extraction) and the NN controller was able to produce the optimal control signals within 1ms on average. Thus, we can achieve the targeted 8FPS as detailed earlier.

\begin{table*}[!tb]
\renewcommand{\arraystretch}{1.3} 
\caption{Comparison between different frameworks}
\label{comparison}
\begin{center}
\begin{tabular}{ c  c  c  c  c  c } 
\toprule
\textbf{Framework} & \textbf{Accuracy (m)} & \textbf{Landing Speed (m/s)} & \textbf{Maximum Altitude (m)} & \textbf{Moving Target} & \textbf{Outdoor}\\
\midrule
Ours & $0.15$ & $0.4$  & $20$  & Yes  & Yes \\ 
Wubben et al. \cite{wubben} & $0.11$ & $0.3$  & $30$  & No  & Yes \\ 
GPS-based & $1-3$ & $0.6$  & $\infty$  & No  & Yes \\ 
Chen et al. \cite{chen} & n/a & $0.23$  & $2.5$  & Yes  & No \\ 
Araar et al. \cite{araar} & $0.13$ & $0.06$  & $0.8$  & Yes  & No \\
Patruno et al. \cite{patruno} & $0.01$ & n/a  & n/a  & No  & Yes \\ 
\bottomrule
\end{tabular}
\end{center}
\end{table*}

\section{Conclusions} \label{conclusions}

We presented an autonomous UAV landing controller that relies only on visual information, to control the UAV towards landing on a moving platform. A target- and environment-agnostic simulator was created and used for training the RL based controller which is responsible for the movements of the UAV. The whole system can achieve speeds of 8FPS, with relatively high accuracy under various environmental conditions, platform speeds and motion patterns. We plan to integrate the controller directly on the UAV by using an embedded board such as NVIDIA\textsuperscript{\textregistered} Jetson or similar. Further, we plan on introducing adaptive speed control for the UAV, based on the platform's speed and motion, resulting in faster landings. Also, GPS assistive data can be used to improve the accuracy of the algorithm in higher altitudes. The accuracy and trustworthiness of the observations provided by the camera can also be quantified in order to improve the control algorithm, as shown in \cite{cheng}, \cite{cheng2}. Finally, part of the future work is to expand our experimental deployments of the system in non-ideal conditions as well.

\end{document}